\pdfoutput=1

\documentclass[11pt]{article}

\usepackage[]{acl}

\usepackage{times}
\usepackage{latexsym}
\usepackage{amsfonts}
\usepackage{amssymb}
\usepackage{pdfpages}
\usepackage{tabularx}
\usepackage{amsmath,bm}

\usepackage[T1]{fontenc}

\usepackage[utf8]{inputenc}

\usepackage{microtype}
\usepackage{amsmath}
\usepackage{diagbox}
\usepackage{graphicx} 
\usepackage{cite}
\usepackage{booktabs}
\usepackage{multirow}
\usepackage{subcaption}
\usepackage{xurl}

\title{Question Difficulty Ranking for Multiple-Choice Reading Comprehension}

\author{Vatsal Raina, Mark Gales \\
  ALTA Institute, University of Cambridge \\
  \texttt{\{vr311,mjfg\}@cam.ac.uk} \\
  }

\begin{document}
\maketitle

\begin{abstract}

Multiple-choice (MC) tests are an efficient method to assess English learners. It is useful for test creators to rank candidate MC questions by difficulty during exam curation. Typically, the difficulty is determined by having human test takers trial the questions in a pretesting stage. However, this is expensive and not scalable. Therefore, we explore automated approaches to rank MC questions by difficulty. However, there is limited data for explicit training of a system for difficulty scores. Hence, we compare task transfer and zero-shot approaches: task transfer adapts level classification and reading comprehension systems for difficulty ranking while zero-shot prompting of instruction finetuned language models contrasts absolute assessment against comparative. It is found that level classification transfers better than reading comprehension. Additionally, zero-shot comparative assessment is more effective at difficulty ranking than the absolute assessment and even the task transfer approaches at question difficulty ranking with a Spearman's correlation of 40.4\%. Combining the systems is observed to further boost the correlation.  

\end{abstract}

\section{Introduction}

With an increasing number of learners of English, there is an increased demand for assessing English reading comprehension (RC) skills. This enables candidates to demonstrate English proficiency, which is often a requirement for business and education institutions \citep{kurz1999review}. 
In the education sector, multiple-choice (MC) tests have repeatedly demonstrated to be efficient in assessing the abilities of candidates' RC skills \citep{alderson2000assessing,moss2001multiple}. However, content creators need to ensure the MC questions are carefully curated to cater to the level of the test takers \citep{burton1990prepare}. In particular, it is important to quantify the difficulty of each MC question. The difficulty of an MC question is influenced by the context paragraph, the specific question and the choice of the distractor options. The difficulty of a question is indicated by the distribution of performance of a cohort of human test takers. An easy question can expect a sharp distribution about the correct answer option while a hard question's probability mass is distributed across the distractor options.

We investigate approaches to automatically rank MC reading comprehension questions by difficulty. Due to a limited number of real datasets with continuous measures of the difficulty of MC questions \citep{settles2020machine}, there have been scant efforts to develop MC question difficulty estimators \footnote{BEA shared task 2024 looks at medical question difficulty. Our focus is on difficulty ranking for language learning.}.
With the recent emergence of the Cambridge Multiple Choice Questions Reading Dataset (CMCQRD) \citep{mullooly2023cambridge, liusie2023analysis} that has difficulty scores for each question, we propose sensible strategies for the underexplored task of question difficulty ranking. Due to the lack of a sufficiently large corpus of MC questions with continuous difficulty scores, the training-based approaches cannot be directly trained for the task of question difficulty estimation. We explore two categories of approaches: task transfer and zero-shot.

\begin{figure*}[t]
    \centering
        \includegraphics[width=0.8\linewidth]{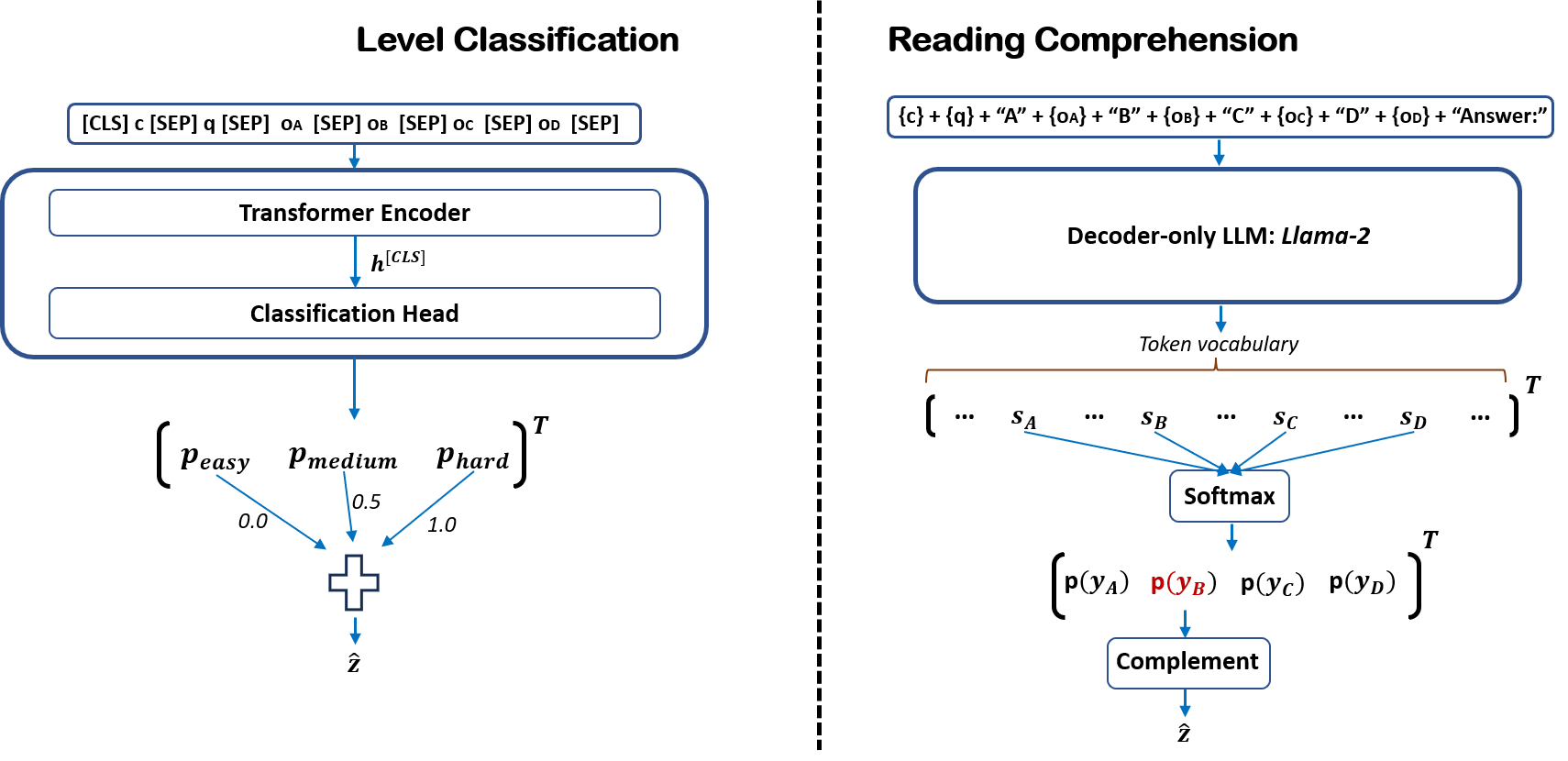}
    \caption{Task transfer for difficulty estimation with context, $c$, question, $q$ and options, $o$.}
    \label{fig:arch_estimation}
\end{figure*}

For task transfer, we adopt two training strategies: 1. train a level classifier system for different difficulty classes (easy, medium, hard) of MC questions on publicly available datasets \citep{pmlr-v101-liang19a} and then map to a continuous score; 2. Train a standard MC reading comprehension system \citep{baradaran2022survey} and manipulate the output distribution as a question difficulty estimate. Both of these approaches rely on supervised data for training.
Additionally, we consider zero-shot approaches which can be applied out-of-the-box for difficulty estimation and hence difficulty ranking. With the universal capabilities of instruction-tuned large language models \citep{minaee2024large} (LLMs), we consider two zero-shot applications: 1. direct prompt for each MC question to return an absolute score of its difficulty; 2. pairwise comparison prompt to request the model to return which question is more difficult between a pair and then convert to an overall ranking across the corpus of questions. Both strategies only rely on black-box API access to instruction-tuned language models.

\section{Related Work}


In the medical domain, \citet{xue2020predicting} and \citet{yaneva2019predicting} explore systems to predict item difficulty for a high-stakes medical exam. In language learning, \citet{huang2017question} propose an explicitly trained convolutional system on the proprietary difficulty score data from IFLYTEK while \citet{beinborn2015candidate} investigate difficulty prediction on cloze-style tests. \citet{loukina2016textual} look at textual properties to predict item difficulty on a proprietary large-scale listening test. We focus on MC question difficulty ranking on publicly available language learning data. 
\citet{liusie2023zero} demonstrated that instruction-finetuned models struggle at absolute assessment of an interpretable metric but give more promising results when used with pairwise comparison. Their work focused on information consistency in summarization. We probe the benefits of pairwise comparison for ranking the difficulty of MC questions.

\section{Difficulty Ranking}
\label{sec:theory}

MC reading comprehension difficulty estimation is a form of the cold-start parameter estimation problem \citep{mccarthy2021jump}. We define the task as follows.
With a corpus of MC questions where each item consists of a context passage, $c$, question, $q$, a set of options, $\{o\}_i$ where $o_j$ denotes the correct answer and $o_{i\neq j}$ denotes the distractors, rank all items by difficulty.

Typically, the automated systems return a difficulty score, $z$, which is used to determine the ranking. Hence, we are only interested in the relative values of $z$, not the absolute. 
We define the difficulty of an MC question according to \citet{mullooly2023cambridge}. It is assumed that there is access to distributions over the options for each question based on human test takers. Following item response theory, the dichotomous Rasch model \citep{andrich1988rasch} is adopted where the Rasch difficulty estimates are cohort independent. Hence, given human cohort distributions across question sets where cohorts of different abilities have attempted different questions, the difficulty score is not influenced by how strong or weak the human candidates are in a given cohort. 
In practice, there is limited data available to train a model directly to predict a difficulty score. Hence, we consider approaches without access to difficulty labelled in-domain training data.

\subsection{Task transfer}

Despite not having access to difficulty scores, we can train systems on alternative complementary tasks. The outputs from such systems can be adapted to generate a continuous difficulty score on an MC question. We explore 2 such tasks.
\newline\newline\noindent
\textbf{Level classification:} Several MC reading comprehension datasets \citep{pmlr-v101-liang19a, lai2017race} are partitioned into different difficulty levels. Hence, we train a classification model to predict the level. Consistent with \citet{raina2022multiple}, Figure \ref{fig:arch_estimation} presents the system. The concatenated inputs are passed through a transformer encoder \citep{vaswani2017attention} and a linear layer to return a probability distribution, $[p_{\text{easy}}, p_{\text{medium}}, p_{\text{hard}}]$ over the levels. The continuous difficulty estimate, $\hat{z}$, is:
\begin{equation}
    \hat{z} = 0.0 \times p_{\text{easy}} + 0.5 \times p_{\text{medium}} + 1.0 \times p_{\text{hard}}
\end{equation}

\noindent \textbf{Reading comprehension:} MC reading comprehension is a standard task where the correct option must be selected for a question on a given context. Following \citet{liusie2023zero, wang2024information}, Figure \ref{fig:arch_estimation} depicts how an instruction finetuned language model, Llama-2 \citep{touvron2023llama2}, is further finetuned for the task. The context, question and answer options are collectively concatenated into a single prompt. The transformer decoder autoregressively is forced to return a single logit distribution over the token vocabulary. Softmax normalizes the logits to get a probability distribution, $[p_{o_\text{A}}, p_{o_\text{B}}, p_{o_\text{C}}, p_{o_\text{D}}]$, over the tokens \textit{A,B,C,D}. The difficulty estimate is taken as the complement of the true class probability.
$
    \hat{z} = 1 - p_{o_\text{j}}.
$

\subsection{Zero-shot}

Since the recently popularized ChatGPT, instruction finetuned large language models \citep{touvron2023llama2, jiang2023mistral, jiang2024mixtral, tunstall2023zephyr, team2024gemma} have demonstrated impressive zero-shot performance across various tasks. They tackle novel instructions without being explicitly trained on such tasks. These models have repeatedly demonstrated to be effective for evaluation \citep{chiang2023closer}. Hence, we consider 2 zero-shot approaches, inspired by \citet{liusie2023zero}, for MC difficulty estimation. 
\newline\newline\noindent
\textbf{Absolute:} For a given MC question, the LLM is prompted to return an absolute score for the difficulty. The model is requested to return a score in the range 1 to 10 where 10 is most difficult. See Appendix \ref{app:prompts} for the exact prompt. For each question, $K$ samples are drawn of the difficulty estimates. The final difficulty score is taken as the mean of the $K$ samples. 
\newline\newline\noindent
\textbf{Comparative:} For a target MC question, $K$ other questions are randomly selected from the corpus. The LLM is pairwise prompted to state which question is more difficult between the target and each of the $K$ other questions. A \textit{win} is defined if the target is deemed more difficult than the other question in the comparison. The difficulty score for the target question is simply the count of its wins. See Appendix \ref{app:prompts} for the prompt.

\section{Experiments}

\subsection{Data}

\begin{table}[htbp!]
\centering
\small 
\begin{tabular}{l|ccc|c}
\toprule
Split & easy & medium & hard & All \\
\midrule 
train & 25,241 & 62,445 & 12,702 & 100,388 \\
valid & 1,436 & 3,451 & 712 & 5,599  \\
test & 1,436 & 3,498 & 708 & 5,642 \\
   \bottomrule
    \end{tabular}
\caption{RACE++ dataset details.}
\label{tab:race}
\end{table}

\noindent For the training, hyperparameter tuning and testing of the level classification and reading comprehension systems, we use RACE++ \citep{pmlr-v101-liang19a}, which is partitioned into three difficulty levels: middle school (easy), high school (medium), and college (hard). Therefore, each question has two labels: the difficulty level and the correct answer. Table \ref{tab:race} summarizes the details.
As a test set, we use CMCQRD \citep{mullooly2023cambridge}. CMCQRD is a small-scale pre-testing stage MC reading comprehension dataset partitioned into levels B1 to C2 on the Common European Framework of Reference for Languages. Each question in this dataset is annotated with a continuous difficulty score. See Table \ref{tab:human} for the counts in each split.

\begin{table}[htbp!]
\centering
\small 
\begin{tabular}{l|cccc|c}
\toprule
& B1 & B2 & C1 & C2 & All \\
\midrule
Count & 140 & 327 & 137 & 54 & 658 \\
Count-dist & 115 & 222 & 72 & 39 & 448 \\
Spearman & 95.1 & 97.2 & 92.3 & 79.5 & 72.0 \\
   \bottomrule
    \end{tabular}
\caption{Human true class probability correlation with difficulty scores on subset of CMCQRD with human distributions.}
\label{tab:human}
\end{table}

\begin{figure}[htbp!]
    \centering
    \includegraphics[width=1.0\linewidth]{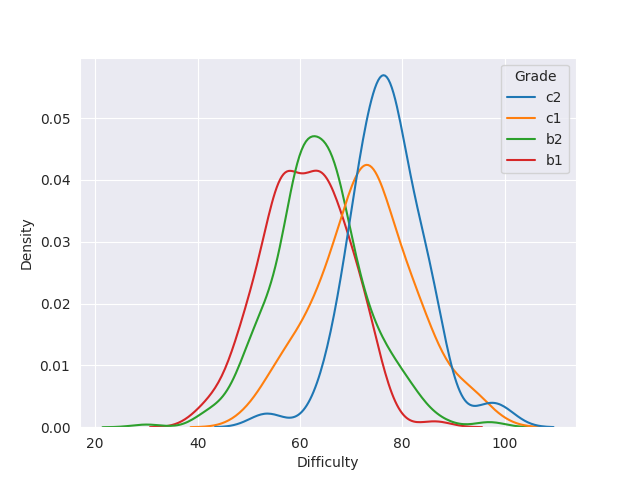}
    \caption{Distribution of difficulty scores in CMCQRD.}
    \label{fig:grade_dist}
\end{figure}

\noindent Figure \ref{fig:grade_dist} presents the distribution of difficulty scores across the grades of CMCQRD. As expected, despite substantial overlap, B1 has the lowest while C2 has the highest mean. 448 questions of the 658 have distributions from the human test takers (denoted count-dist). Hence, it is interesting to check the correlation between the probability associated with the true class and the attributed difficulty score. Table \ref{tab:human} calculates the ranking agreement using Spearman's correlation on the subset of CMCQRD with human distributions. The correlation is the weakest for C2, which is supported by Figure \ref{fig:grade_dist} as C2 has the narrowest shape. The correlation is substantially weaker when all grades are combined together, reflecting that cohorts of different abilities are taking the tests for each grade (see the definition of difficulty in Section \ref{sec:theory}). 

\subsection{Model details}
\label{sec:model}

For the level classifier, an ensemble of 3 models are trained on the RACE++ train split to predict easy, medium or hard. ELECTRA-base \citep{clark2020electra} is the backbone pre-trained model.
For the reading comprehension system, Llama-2 is finetuned to select the correct answer on the train split of RACE++.
See Appendix \ref{sec:hyp} for hyperparameters.
For the zero-shot approaches, ChatGPT3.5 is used. See Appendix \ref{app:open} for alternative models.


\section{Results}
\label{sec:results}

\begin{table}[htbp!]
\centering
\small 
\begin{tabular}{ll|c}
\toprule
Approach &  &  Spearman \\
\midrule
Task transfer & Level classification (LC) & 38.7 \\
& Reading comprehension & 30.8 \\
\midrule
Zero-shot & Absolute & $ 18.3_{\pm 0.3}$ \\
& Comparative & $40.4_{\pm 0.6}$ \\
\midrule
Combined & LC $\oplus$ Comparative & $43.7_{\pm 0.6}$ \\
   \bottomrule
    \end{tabular}
\caption{Difficulty score Spearman's rank correlations. }
\label{tab:comp_eval}
\end{table}

\noindent Table \ref{tab:comp_eval} presents the Spearman's correlation between the systems' estimates and the true difficulty scores in CMCQRD. The task transfer approaches show a positive correlation to rank MC questions. Particularly the level classification system hits 38.7\%.  
See Appendix \ref{app:level} and \ref{app:read} for the performance of the systems. 

\begin{figure}[htbp!]
     \centering  \includegraphics[width=1.0\columnwidth]{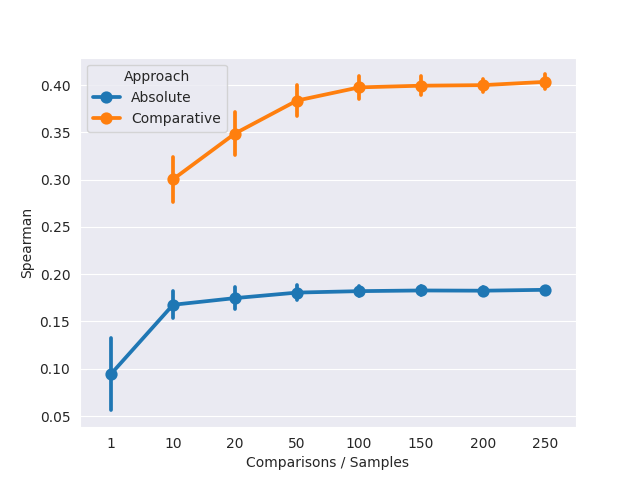}
        \caption{Correlation using zeros-shot approaches.}
        \label{fig:pair}
\end{figure}

For zero-shot, Table \ref{tab:comp_eval} presents the correlation with absolute and comparative assessment. The absolute system difficulty estimate is determined by taking the average score from $K=250$ samples from ChatGPT. For the comparative system, $K=250$ comparisons are made for each question and the difficulty estimate is taken as the number of wins out of 250. The standard deviation is calculated over 30 draws. The instruction-tuned models are more effective at ranking question difficulty if used for comparative rather than absolute assessment. It is further interesting that the zero-shot comparative approach is able to out-compete even the task transfer level classification approach. ChatGPT was observed to have a 23.8\% bias to select the first item over the second in the prompt during comparison. Figure \ref{fig:pair} explores how the performance of the absolute and comparative systems vary with $K$. We observe that as $K$ increases, the performance plateaus with decreasing variance. 

Finally, Table \ref{tab:comp_eval} presents the correlation by combining the level classification and zero-shot comparative approaches by averaging ranks. We observe a significant performance jump to 43.7, suggesting the two approaches have complementary behaviour in ranking the difficulty of MC questions.

\section{Conclusions}

This work explores task transfer and zero-shot approaches for ranking MC questions by difficulty on CMCQRD that has continuous difficulty scores. A RACE++ trained level classification system transfers better to the task of difficulty ranking than a reading comprehension system. It is further seen that zero-shot comparative difficulty ranking by instruction finetuned models outperforms their absolute estimation of difficulty. Finally, combining the level classification and comparative systems observes a further performance boost.  

\section{Acknowledgements}
This research is partially funded by the EPSRC (The Engineering
and Physical Sciences Research Council)
Doctoral Training Partnership (DTP) PhD studentship
and supported by Cambridge University Press \& Assessment (CUP\&A), a
department of The Chancellor, Masters, and Scholars
of the University of Cambridge.

\section{Limitations}

The approaches in this work, due to limited public data availability of continuous difficulty scores in language learning, rely on a single corpus, CMCQRD, for ranking MC questions by difficulty.

\section{Ethics statement}

There are no ethical concerns with this work.

\bibliographystyle{acl_natbib}
\bibliography{custom}

\appendix

\newpage
.
\newpage

\section{Prompts}
\label{app:prompts}

This section states the prompts for the zero-shot approaches for difficulty estimation.
\newline\newline\noindent
\textbf{Absolute}: `` \{context\}$\backslash$n$ \backslash$n \{question\}$\backslash$n A) \{option\_A\}$\backslash$n B) \{option\_B\}$\backslash$n C) \{option\_C\}$\backslash$n D) \{option\_D\}$\backslash$n  $\backslash$n Provide a score between 1 and 10 that measures the difficulty of the question. Return only a single score. ''
\newline\newline\noindent
\textbf{Comparative}: `` 1: $\backslash$n $\backslash$ n \{context\_1\}$\backslash$n $\backslash$n \{question\_1\}$\backslash$n A) \{option\_A\_1\}$\backslash$n B) \{option\_B\_1\}$\backslash$n C) \{option\_C\_1\}$\backslash$n D) \{option\_D\_1\}$\backslash$n $\backslash$n $\backslash$n 2: $\backslash$n $\backslash$n \{context\_2\}$\backslash$n $\backslash$n \{question\_2\} $\backslash$n A) \{option\_A\_2\}$\backslash$n B) \{option\_B\_2\}$\backslash$n C) \{option\_C\_2\}$\backslash$n D) \{option\_D\_2\}$\backslash$n $\backslash$n $\backslash$n  Which reading comprehension question is more difficult, 1 or 2? Return only 1 or 2. ''

\section{Hyperparameters}
\label{sec:hyp}

\textbf{Level classification system}: With grid search hyperparameter tuning on the valid split, the final values are: AdamW optimizer, batch size 3, learning rate of 2e-5, max epochs of 3 with total training taking 3 hours on NVIDIA V100.
\newline\newline\noindent
\textbf{Reading comprehension system}: We use parameter efficient finetuning (due to compute limits) with quantized low rank adapters (QLoRA) \citep{dettmers2023qlora}. Final hyperparamters include: learning rate of 1e-4, batch size of 4, lora rank of 8, lora $\alpha$ of 16 and dropout 0.1. Training takes 7 hours for 1 epoch on NVIDIA A100.

\section{Level classification system performance}
\label{app:level}

This section reports the performance of the level classification system on the RACE++ validation and test sets. Alongside accuracy, the performance is reported using F1 to account for the imbalance between the \textit{easy}, \textit{medium} and \textit{hard} classes. The trained system achieved 87.2\% accuracy and 83.7\% F1.

\section{Reading comprehension system performance}
\label{app:read}

This section reports the performance of the reading comprehension system on the RACE++ and CMCQRD test sets. The overall accuracy is reported of identifying the correct answer option where a random system would achieve 25\%. The system achieves 86.0\% on RACE++ and 79.9\% on CMCQRD.

\section{Open-source LLM results}
\label{app:open}

The main paper results investigate pairwise comparative approaches by prompting ChatGPT. Here, we additionally report the results from a small open-source instruction finetuned LLM, Mistral-7B. Using the smaller model, with 100 pairwise comparisons, the Spearman's rank correlation is 17.5\%. The lower performance compared to ChatGPT suggests the smaller model struggles to make sensible comparisons for a task with this challenge level.

\section{Licenses}

For CMCQRD, the license\footnote{Available at: \url{https://englishlanguageitutoring.com/datasets/cambridge-multiple-choice-questions-reading-dataset}} states the licensed dataset for non-commercial research and educational purposes only.

\end{document}